\documentclass[letterpaper]{article} %
\usepackage{aaai25}  %
\usepackage{times}  %
\usepackage{helvet}  %
\usepackage{courier}  %
\usepackage[hyphens]{url}  %
\usepackage{graphicx} %
\usepackage{natbib}  %
\usepackage{caption} %
\usepackage{algorithm}
\usepackage{algorithmic}

\usepackage{tabularx}
\usepackage{longtable}
\usepackage{natbib}
\usepackage{bibentry}

\DeclareUnicodeCharacter{0301}{\'{e}}

\usepackage{adjustbox}

\usepackage{physics}
\usepackage{amsmath}
\usepackage{tikz}
\usepackage{mathdots}
\usepackage{yhmath}
\usepackage{cancel}
\usepackage{color}
\usepackage{siunitx}
\usepackage{array}
\usepackage{multirow}
\usepackage{amssymb}
\usepackage{gensymb}
\usepackage{tabularx}
\usepackage{extarrows}
\usepackage{booktabs}
\usetikzlibrary{fadings}
\usetikzlibrary{patterns}
\usetikzlibrary{shadows.blur}
\usetikzlibrary{shapes}

\usepackage{newfloat}
\usepackage{listings}
\DeclareCaptionStyle{ruled}{labelfont=normalfont,labelsep=colon,strut=off} %
\floatstyle{ruled}
\newfloat{listing}{tb}{lst}{}
\floatname{listing}{Listing}
\usepackage{hyperref}%

\title{What Do Machine Learning Researchers Mean by ``Reproducible''?}
\author{
    Edward Raff\textsuperscript{\rm 1,2},
    Michel Benaroch\textsuperscript{\rm 3},
    Sagar Samtani\textsuperscript{\rm 4},
    Andrew L. Farris\textsuperscript{\rm 1}
}
\affiliations{
    \textsuperscript{\rm 1}Booz Allen Hamilton,
    \textsuperscript{\rm 2}University of Maryland, Baltimore County,
    \textsuperscript{\rm 3}Syracuse University,
    \textsuperscript{\rm 4}Indiana University
}

\begin{document}

\maketitle

\begin{abstract}
The concern that Artificial Intelligence (AI) and Machine Learning (ML) are entering a ``reproducibility crisis'' has spurred significant research in the past few years. Yet with each paper, it is often unclear what someone means by ``reproducibility''. 
Our work attempts to clarify the scope of ``reproducibility'' as displayed by the community at large. In doing so, we propose to refine the research to eight general topic areas. In this light, we see that each of these areas contains many works that do not advertise themselves as being about ``reproducibility'', in part because they go back decades before the matter came to broader attention. 
\end{abstract}

\section{Introduction}

The Artificial Intelligence (AI) and Machine Learning (ML) communities are increasingly concerned with the ``reproducibility'' of their fields. This has come on the heels of a reproducibility crisis noted in many others. We will refer to this overarching concern, that the science of research is being done with some error rate, as a generic \textit{scientific rigor} concern. This concern is justified, and
it is increasingly challenging to evaluate the state of research around scientific rigor due to confused and incompatible usage of the same few terms like ``reproducibility'' \citep{Plesser2018}.

Due to confusing and often inconsistently used terminology in the
literature, it is challenging to understand precisely what issues of
scientific rigor the community is tackling. In light of these issues, 
we propose a new formulation of
current scientific rigor research by surveying the current articles by the
topics they cover. In doing so, we observe that many historical works tackled these very issues -- with different motivations and no particular thematic name like ``reproducibility'' as it was not an urgent concern at the time. 

In this article, we will expand the ACM's proposed terminology of
Repeatability, Reproducibility, and Replicability which we find useful,
although still insufficient to capture the breadth of work done to
date. Our contribution classified current AI / ML research in scientific rigor into eight aspects we label as \emph{repeatability},
\emph{reproducibility, replicability}, \emph{adaptability}, \emph{model
selection}, \emph{label/data quality}, \emph{meta \& incentive}, and
\emph{maintainability}. These eight aspects are defined in \autoref{tbl:themes}. We
propose these aspects based on our review of 101 papers published since
2017 and reflect the focus of the community at large. \autoref{tbl:themes} also shows
for each aspect the proportion of papers focused primarily on that
aspect (though many papers touch on multiple aspects). 

\begin{table*}[!h]
\caption{Eight primary topics that have been collectively described as ``reproducibility'' in the literature, determined by our manual review. The first three are based on the ACM's guidelines, and the rest are informed by surveying and categorizing the themes of existing literature. } \label{tbl:themes}
\begin{tabularx}{\textwidth}{lXr}
\toprule
\multicolumn{1}{c}{Topics} & \multicolumn{1}{c}{Main Concern}                                                                                                                                         & \multicolumn{1}{c}{\% of papers} \\ \midrule
Repeatability              & Can the results be obtained by the original authors using their original code and data.                                                                                  & 12.9                             \\
Reproducibility            & Can a different team obtain the results using the code and data provided by the original authors                                                                         & 16.8                             \\
Replicability              & Can a different team, using different code and/or data, obtain the same results or results congruent with the original publication.                                      & 15.8                              \\
Adaptability               & Can the original authors using the original code obtain qualitatively similar results on new/different data.                                                             & 4.0                              \\
Model Selection            & Given a set of two or more models, what process can be used to meaningfully and reliably determine which model to select for use.                                        & 19.8                             \\
Label/Data Quality         & Given a process for labeling data, how can we ensure that the process results in meaningfully same labels over time and that the process of labeling has minimal errors. & 4.0                              \\
Meta \& Incentive          & What are the motivators, or lack thereof, for scientific rigor.                                                                                                          & 13.9                             \\
Maintainability            & What are the issues and remediations in running the same AI/ML solution as the people, code, and data are all altered in their nature over time.                         & 12.9                             \\ \bottomrule
\end{tabularx}
\end{table*}

The rest of this article is organized as follows.
We will summarize the eight main topic areas of scientific rigor in \autoref{sec:details}, with sub-areas included based on our literature review. Based on this survey of the literature, we propose relationships for how these rigors interact in \autoref{sec:connections}, which we find informative as a macro-level picture of the scope of scientific rigor. 
Finally, we conclude in \autoref{sec:conclusions}.

\section{The Current Scope of Work} \label{sec:details}

Our literature survey identifies at least eight primary aspects of
scientific rigor studied in the AI/ML literature. Each major sub-section will repeat one of the eight rigors defined in \autoref{tbl:themes}, and include further delineation for nuanced sub-categories that are present or noteworthy in the literature. A key criterion for being included in \autoref{tbl:themes} is that the paper must self-identify itself as being about ``repeatability, reproducibility, or replicability'' since those are the three preexisting terminologies used (interchangeably) in the prior literature. These delimitations reflect the current scope of what researchers actively consider ``reproducibility'' consider worthy of study and effort. 
As our bibliography will show though, many more papers exist in these topical areas that were published before 2017, and thus before the AI/ML communities started to put renewed effort into the issue of scientific rigor. We include such articles in the discussion of each section to establish the full scope of available work and to connect the current reproducibility-themed motivation to its historical precedents, as the historical literature is often unknown to existing researchers on this topic. 

Before we detail these aspects, it is worth noting that many existing articles are best summarized as opinion pieces with varying degrees of formalization of their arguments. Most of these articles propose strategies or arguments on how to obtain ``reproducibility'', without evidence of effect \citep{Gundersen2018,matsui_mlops_2022,Publio2018,Tatman2018,sculley2018winner,Vollmerl6927,Drummond2009,Drummond2018,Raff2022a,lin_building_2022}. 
The contents of our article are focused on works that study issues, incentives, or interventions to rigor issues in AI/ML --- and thus go beyond opinion or thought pieces on the topic, of which there are many. These thought pieces are valuable in spring motivation and growth in the field, but most are disconnected from the long literature on the topic, so we prefer not to focus on their opinions which may well change with new literature\footnote{Indeed, our own understanding have evolved in the discovery of the wide and deep literature on scientific rigor.}. 

Since there is no canonically accepted ``home'' for AI/ML papers on reproducibility, we find that such published literature is scattered across various subfields and specialized conferences. In many cases, we find common themes in the nature of issues that occur across fields and domains, and in some aspects, the literature on issues impacting scientific rigor directly goes back to the 1990s. Our categorization is based on a review of all literature we are aware of that tackles scientific rigor issues, even if they did not use terms like ``reproducible'' as they often pre-date the larger academic concern itself. One recommendation that we would put forth for \textit{all major AI/ML conferences is to create a track for scientific rigor studying all eight proposed rigor topics to further incentivize and organize this important work}.

\subsection{Repeatability}

Repeatability concerns the authors who obtain the same results using the original source code and data. 
Interesting questions in repeatability include how to develop code and
systems that make it easy for the developer to keep track of how they
came to their experimental results from an experimental design
perspective \citep{Gardner2018,Paganini2020}.
In Human-Computer Interaction (HCI) research, there has been significant
study on the iterative development nature of computational notebooks
(e.g., Jupyter) that are widely used in AI/ML development processes.
These notebooks can be prone to many subtle code errors/issues due to
their fluidity and out-of-order execution. Enhanced tools can ensure the
exact execution sequence to generate a result \citep{head_managing_2019,kery_interactions_2018,10.1145/3641525.3663622}. Many simple factors, like using a random-number seed (i.e., for a pseudo-random-number generator (PRNG)), are important for obtaining instantaneous repeatability. Furthermore, many mathematical operations are not guaranteed to produce identical numerical results due to floating point errors and differences in numerical stability of different implementations and hardware~\cite{6877351,NEURIPS2023_af076c3b}. 

Other factors, such as software version conflicts, are often thought to be factors of repeatability but often lead to conflation. For example, does capturing software versions via a container system lead to repeatability or reproducibility? We argue that it would be reproducibility as a higher-level concept in our categorization, which we will detail further in \autoref{sec:connections}. A second distinction we make is that of instantaneous repeatability vs. repeatability over time. In this immediate section, we consider instantaneous repeatability, where the question is how to ensure repeatable results as the software/algorithm is being developed, and we find that there is surprisingly little beyond the work noted in the prior paragraph. When time is added as a factor, we consider this to be distinguishable as the maintainability rigor that we will detail in \autoref{sec:maintainability}.

\subsection{Reproducibility}

Reproducibility alters repeatability by requiring that a different individual/team be able to produce the same results using the original source code and data. This is a high focus of the AI/ML community and incentivization of Open Source Software (OSS) by major conferences and paper submission questionnaires/guidelines. 
Current work can be divided into those that explore surface-level issues
such as unquantified proposals or exact procedure reproductions, vs
those that attempt to quantify or better understand why a reproduction
does(not) work.

\subsubsection{Surface Reproducibility}

Surface-level studies of reproducibility report on the scale of the
reproducibility challenge without examining whether their attempts at
improving reproducibility work. The only large-scale study we are aware of found that 74\%
of the code released by the broader scientific community (beyond AI/ML) ran
without issue \citep{trisovic_large-scale_2022}. Toward remediating this in machine
learning, many have proposed techniques like Docker to try and capture
the exact conditions to re-run the experiments \citep{Forde2018,Forde2018ReproducingML}.
\cite{8621874} looked at enhancing reproducibility by standardizing data access and execution environments for MOOCs. However, the project appears to be abandoned and stresses the importance of repeatability/reproducibility over time, which we note forms the aspect of maintainability we discuss later in \autoref{sec:maintainability}. 

\subsubsection{Reproducibility In Depth}

A major factor in Reproducibility, and the discovery of non-reproducible work, is errors in the original comparisons being made. There are cases where reproducibility may be strictly achievable but meaningless due to an error in the fundamental approach being taken or experimental setup. In a seminal example of metric learning, it was found that papers had multiple changes occurring simultaneously in comparison to prior baselines (new layers like Batch-Norm, optimizers, etc.) beyond just the proposed metric learning changes, which produced misleadingly large effect sizes \citep{Musgrave2020}. In general, many other works have identified similar issues nuanced to the subdomain being studied \citep{lu_coreset_2023,Liu2020e,chen-etal-2022-reproducibility,ito-etal-2023-challenges}. More serious instances have determined a subfield of research being constructed around unsound methodologies~\cite{Lin_Liu_Chen_Hsu_Wu_Tsai_Lin_2022,KAPOOR2023100804,10.1145/3514094.3534196,raff2023reproducibility}.

Thematically similar to \cite{Musgrave2020} are the multiple realizations of insufficient baseline evaluation that have occurred in many works since. Such work includes studies that use similar baseline errors/lack of adjustment \citep{rao_where_2022}, studies that expand the set of baselines against an overly broad prior conclusion \citep{huang_state_2022,wang_inspection_2022}, and studies that demonstrate that decades-old methods are still competitive when given the chance to run on larger modern datasets \citep{liu_another_2022}. Another example is the effectiveness of linear models in natural language processing tasks, which are orders of magnitude faster and capable of comparable results~\cite{lin-etal-2023-linear}. A unique aspect shown by
\cite{chen-etal-2018-best} is that many improvements prescribed to one family of algorithms are actually applicable to prior approaches and would perform just as well using an ``out-of-date'' method. They showed this by applying improvements from seq2seq modeling to Recurrent Neural Networks and found that the improvements were still effective, allowing a Pareto improvement in combined approaches.

\subsection{Replicability}

Replicability concerns the ability of a different person/team to produce qualitatively similar results from the original article by writing their own code and potentially different data. 
The aspect of replicability is highly understudied, likely due to the challenges this aspect presents. 
Replicability can be subdivided into empirical replicability and theoretical replicability. 

\subsubsection{Empirical Replicability}
Empirical Replicability requires re-implementing a target method's code from scratch, which is a labor-intensive process. Notable work in this direction was done by \citep{Raff2019_quantify_repro}, who attempted to reimplement 255 papers, and computed features to quantify what properties correlated with a replicable paper. Smaller scale replications have also been performed \citep{belz-etal-2022-2022}, including a volunteer effort by ReproducedPapers.org collecting some (most are reproduction attempts) Replicability attempts \citep{yildiz_reproducedpapersorg_2021} based on which a thorough study in IR has been performed \citep{wang_inspection_2022}. \cite{chen-etal-2022-even,Ganesan2021} identified issues with a specific common baseline method XML-CNN in multi-label learning. Famously, \cite{10.5555/3504035.3504427} replicated recent reinforcement learning results and discovered various aspects, such as the seed and scale of rewards, that significantly altered the perception of improvement. 

\cite{pmlr-v68-johnson17a} Replicate studies in mortality prediction in a healthcare context, highlighting the difficulty of producing comparable results when replication also requires collecting new data of the same intrinsic nature (that is, patient data in this context). Textual descriptions presented in the original studies were found to be insufficient for collecting new data that would replicate. \cite{pmlr-v85-hegselmann18a} extended this observation by showing how to produce replicable data collection schemes for survival analysis against medical repositories such as SEER. 

We are not aware of other work within AI/ML on empirical replicability. This state of affairs is common to other (relatively) code-free disciplines such as medicine \citep{Ioannidis2005} economics \citep{RePEc:pra:mprapa:75461} social sciences \citep{Camerer2018}. In these disciplines, replication studies are necessary and representative because they are the least costly way to evaluate a result. Other aspects of scientific rigor have a significantly lower barrier to entry, largely because AI/ML has a large open-source culture.

\subsubsection{Theoretical Replicability}

More recent work has advanced a theoretical definition of replicability in terms of constraints on the output distribution as a function of the input distribution. ~\cite{10.1145/3519935.3519973,pmlr-v202-kalavasis23a,10.1145/3564246.3585246} have developed much of this foundation by showing various desirable statistical properties such as that Total Variation (TV) between outputs drawn from the same distribution using the same algorithm (i.e., a congruent definition of replicability) is equivalent to results in approximate differential privacy and robust statistics. This idea has since been expanded to bandits~\cite{esfandiari2023replicable}, optimization~\cite{zhang2023optimal}, clustering \cite{esfandiari2023replicable}, and reinforcement learning (at an exponential increase in runtime)~\cite{karbasi2023replicability}. 

Lastly, a unique approach to the question of replicability was studied by \citep{Ahn2022}, which focuses on the difference between the computational precision of floating points and the underlying symbolic math. From this perspective, they are able to suggest conditions about what statements can be rigorously tested and concluded about the math based on floating-point errors that would accumulate and cause issues otherwise. 

\subsection{Model Selection}

Model selection deals with the common task of AI/ML papers: given two competing methods (one of which may be the paper's own proposal), how do we conclude which method is better?
As the AI/ML literature has advanced significantly through the presentation of empirically ``better'' algorithms, it is not surprising that most historical and current work has focused on the question of model selection. This includes how to pick and evaluate criteria to decide ``better'', how to build benchmarks for a problem, and the process to determine ``better'' given criteria in a statistically sound way. There are also multiple resurgences of this issue as ML is incorporated into other fields, and comparisons that may be invalid in a new field occur as both communities begin to merge and discover what is/is not acceptable ~\cite{9847318}. 

\subsubsection{Evaluation Criteria \& Methodology}

Before selecting the ``better'' of one or more methods, it is necessary first to determine how the quality of a method is determined. The scope of evaluation metrics and scores is larger than that of scientific rigor, and this article is concerned with cases where an invalid or errant procedure was identified and remediated. The literature in this direction is old, starting in the late 1990s on the various pros and cons of metrics like Area Under the Curve (AUC) for evaluation \citep{Bradley1997,hand_measuring_2009,lobo_auc_2008}. Likewise, work has addressed issues in scoring from leaderboards \citep{icml2015_blum15}, and subtle issues in using cross-validation to produce test scores \citep{Varma2006,Bergmeir:2018:NVC:3178572.3178665,VAROQUAUX201868,Bates2021,10.1145/3641525.3663618}. Niche examples of the evaluation concern also exist. For example, three decades of malware detection performed subtle train/test leakage by adjusting for a target false-positive rate incorrectly \citep{Nguyen2021} and time series anomaly detection scores being overly generous to ``near hits'' \citep{Kim2022}.

\subsubsection{Building Problem-Specific Benchmark Suites}

It is becoming increasingly popular to build benchmarks of multiple datasets, pre-prepared evaluation code, and methodology for specific problem domains \citep{mlsys2020_73,Eggensperger2021,10.1145/3383313.3412489,saul2024is,liu2024assemblage,Ordun2021a,Kebe2021a}. Such a benchmark construction is popular, although it has yet to evolve into a science of how to build benchmarks, with limited study at a macro level \citep{koch_reduced_2021}. Some domains may require additional thought to how methods are compared, especially when they are measuring a non-stationary objective like human preferences in Information Retreival~\citep{10.1145/3641525.3663619}.

\subsubsection{Selection Determination}

Much of the ML literature presents raw results and makes a nonscientifically rigorous statement of being ``better'' by some metric (i.e., \textbf{bold} numbers in a table are better, and our method has more bold numbers in the table). There are two approaches to developing improved comparisons.

One is to devise better statistical tests to compare two methods when a single test set is available, first seriously studied by \citet{Dietterich:1998:AST:303222.303237} with many follow-up works shortly after \citep{Alpaydin:1999:CTC:339993.339999,bouckaert_choosing_2003,bouckaert_evaluating_2004}. Different perspectives on this include using one test run to make a conclusion \citep{dror-etal-2019-deep}, or including sources of variation in model performance (e.g., hyperparameter values) and comparing the distribution of model results \citep{pmlr-v97-bouthillier19a,Bouthillier2021,Cooper2021}. Others have introduced computational budget for training and parameter tuning as a conditional factor that impacts the conclusion of ``best'' \citep{Dodge2019}.

The second option is to use multiple datasets to perform a single test of whether one algorithm is better than another \citep{guerrero_vazquez_repeated_2001,hull_information_1994,pizarro_multiple_2002}. The use of a nonparametric Wilcoxon test has been found to be effective in multiple studies \citep{Demsar:2006:SCC:1248547.1248548,JMLR:v17:benavoli16a}. \cite{dror-etal-2017-replicability} extended this to make a conclusion about how many datasets and which one method performs better. Other recent work has proposed using meta-analysis methods to draw conclusions about a single method tested under multiple conditions \citep{Soboroff:2018:MRE:3269206.3271719}. Notably, work using multiple datasets to make decisions based on a single evaluation metric implicitly contributes to the Adaptability question, which we explore next. Interestingly, we note the field of programming languages has also proposed quantile regression as a better method of analyzing results ~\cite{10.1145/2499368.2451140}.

\subsection{Adaptability and its Second-Class Status}

Adaptability is the study of a different person/team, using the original code but applying it to their own and different data. Very little work on scientific rigor in AI/ML focuses on Adaptability. To be clear, many prior works have studied the question of generalization in machine learning, of which there is recent evolution due to the advance of deep learning \citep{Zhang2017}. However, generalization assumes some form of intrinsic relationship (usually I.I.D.) between the training and testing distribution. Under Adaptability, there is no direct train/test split to compare. Instead, it is a question of the methodology's effectiveness on an entirely different statistical distribution at training and test time. Thus, our concern is more focused on the practical, real-world issues that enable or inhibit a \textit{method} to generalize. Our contention is the lack of study on adaptability is one of the most glaring shortfalls in the current scientific rigor literature, with significant room for researchers to define and develop new ways of studying the problem\footnote{Anecdotally, we have had significant trouble publishing work attempting to tackle adaptability problems in other sub-domains when we were trying to use it for real-world needs~\citep{Raff2023ers}.}. 

The work we have found can broadly be described as including adaptability to new datasets or specialized subsets to better understand the overall behavior and utility of a set of algorithms \citep{Marchesin2020,rahmani_experiments_2022}. The other work that tackles adaptability is from an HCI perspective in validating a method's utility as population preferences evolve \citep{roy_users_2022}.

Though it has not been presented as a part of the literature on scientific rigor, considerable effort in the Adaptability question has been advanced by Decision Tree-based literature. In particular, the long-standing effectiveness of tree ensembles has led to numerous studies investigating the persistent efficacy of tree ensembles \citep{grinsztajn_why_2022,JMLR:v17:15-374,bagnall2020rotation}. Despite little work on the adaptability question, we note that many works in Model Selection make use of the adaptability argument as a component of their study or an otherwise latent concern. 

One way that others could seek to understand adaptability is to see if they adapt to crossing a small ``chasm'' of change in the problem. This can be done by taking methods developed in one context and applying them to a highly related problem (ideally with a minor to no modification necessary). Two prior works that we are aware of have demonstrated surprising failures of methods that fail to cross these small chasms. \cite{Riquelme_Tucker_Snoek_2018} studied extensions of Thompson Sampling for reinforcement learning that work well in supervised settings, but a modest adaption to sequential decision making causes simple Thompson Sampling to outperform the various previous improvements. 
\cite{liu-etal-2021-parameter} found that the original hyperparameters for a multi-label prediction algorithm were kept when the method was adapted to a new task. Subsequent works compared to this original parameterized version rather than re-tuning to the new task. When properly accounted for, all subsequent methods failed to improve on the original method.

\subsection{Label \& Data Quality}

Label \& Data Quality is focused on the reliability of data and label acquisition, error rates, and working to understand how they occur, detect them, or work around them. The distinction we make from research in inferring a single label from labelers is that scientific rigor is concerned with the process of how labels are collected, defined, and have impacted research conclusions (e.g., inferring a 99\% accurate model when labels have a 5\% noise level would imply a failure in process).
Many works today identify these issues long after dataset construction, in part due to the high accuracies now being achieved, making the errors more pronounced. 
For example, the process for deriving labels of ImageNet had rules incongruent with the nature of the data (e.g. assuming that only one class is present) and error-prone steps in the labeling pipeline \citep{Beyer2020}. Label quality issues also include leakage from the train / test set \citep{Barz2019}.

Some of the most insightful research results have come from replicating dataset construction and labeling processes for prior datasets and then characterizing and discovering why differences in results occur. This includes detection cases where recreation is implicitly made more challenging than the original dataset \citep{pmlr-v119-engstrom20a}. 
Although there is a long history of research inferring a single correct label from multiple labelers \citep{whitehill_whose_2009,lin_relabel_2014,NIPS2016_6523,yoshimura_quality_2017,Ratner2020}, this literature is generally not framed as a scientific rigor issue. While these methods have been utilized in work from a rigor perspective \citep{Beyer2020}, we are not aware of work that bridges a longitudinal study of the replicability of these various label inference procedures. 

\subsection{Meta and Incentives}

Very few papers have studied incentives for scientific rigor. The study of the scientific process itself is often termed metascience and, when applied to AI/ML research, would fall into this category.  Such research could include basic studies of incentives, drivers of scientific rigor, and surveys across various AI/ML research domains. 
Sample studies focused on drivers such as the rate of data and code sharing in computational linguistics \citep{wieling_squib_2018} and the use of statistical testing \citep{dror-etal-2018-hitchhikers}. Related work has found that code sharing and replica research are correlated with higher citations \citep{Raff2022,10.1145/3641525.3663628}, although most meta-studies have looked at the rate of code sharing in their subdisciplines~\cite{McDermott_Wang_Marinsek_Ranganath_Foschini_Ghassemi_2021,10.1145/3576915.3623130,arvan-etal-2022-reproducibility-code,arvan23_interspeech,10.1145/3596519}. A unique aspect of code availability is studied by ~\cite{storks-etal-2023-nlp}, who perform a user study with students on the time and difficulty factors for students to reproduce the results of three NLP papers. Another study focuses on how evaluation and comparison practices evolve throughout the Machine Translation community \citep{marie-etal-2021-scientific}. The last work we are aware of challenged the treatment of replicability as a binary ``yes/no'' question and instead suggested a survival model, where replicability is a function of time/effort \citep{Raff2020c} and quantifying a reproducibility score~\citep{belz-etal-2022-quantified}.

\subsection{Maintainability} \label{sec:maintainability}

Maintainability is similar to Repeatability, in that we are concerned with producing the same results with the original authors (though new users could also occur) using the original code and data. The key difference that distinguishes maintainability is that time is a factor, as the ability to repeat results degrades over time as nuances of labels~\citep{Inel2023} or dependency versions change~\citep{Connolly2023Software}\footnote{In software development, this notion is often termed ``bit rot''. }. Maintainability can also deal with the code itself changing over time. 
The focus on the aspect of maintainability within AI/ML was started by the seminal work of \citep{10.5555/2969442.2969519}. A key area of maintainability deals with adapting known ``code smells'' while considering ML-specific concerns and factors that practitioner surveys consider most important \citep{gesi_code_2022}. Another key area of maintainability is the quality of the results as the code itself changes. It is well known that scientific algorithms may produce different results by different (but supposedly equivalent) implementations \citep{Hatton1993}. Multiple studies have found that AI/ML is no exception to this history, with large and statistically significant changes in accuracy when using allegedly equivalent algorithms and changing just the implementation or the runtime platform (e.g. GPU hardware) \citep{coakley_examining_2022,gundersen_machine_2022,10.1145/3324884.3416545,Zhuang2021}. 
\cite{pmlr-v209-zhou23a} found that in many medical time series tasks, it may be beneficial to train on all historical data in some cases vs. training a sliding window of recent data. They also looked at models that experienced ``shocks'' of sudden degradation in time.

The study of maintainability is surprisingly minimal in our community despite the rapid adoption, abandonment, and evolution of frameworks used within the field. Torch, Tensorflow, JAX, Theano, and many more frameworks have come and gone through major revisions over time. These changes and re-implementation of algorithms are fertile ground for maintenance issues and, thus, their study, which directly impacts researchers and the developers of these frameworks. Studying how to build maintinable code in AI/ML is still nascent~\cite{10.1145/3641525.3663624,papi-etal-2024-good}

\section{Connections between Rigor Types} \label{sec:connections}

\begin{figure}[!h]
    \centering
    \adjustbox{max width=0.95\columnwidth}{%
    \tikzset{every picture/.style={line width=0.75pt}} %

\begin{tikzpicture}[x=0.75pt,y=0.75pt,yscale=-1,xscale=1]
\draw   (10,180) .. controls (10,174.48) and (14.48,170) .. (20,170) -- (120,170) .. controls (125.52,170) and (130,174.48) .. (130,180) -- (130,210) .. controls (130,215.52) and (125.52,220) .. (120,220) -- (20,220) .. controls (14.48,220) and (10,215.52) .. (10,210) -- cycle ;
\draw   (310,180) .. controls (310,174.48) and (314.48,170) .. (320,170) -- (420,170) .. controls (425.52,170) and (430,174.48) .. (430,180) -- (430,210) .. controls (430,215.52) and (425.52,220) .. (420,220) -- (320,220) .. controls (314.48,220) and (310,215.52) .. (310,210) -- cycle ;
\draw   (160,180) .. controls (160,174.48) and (164.48,170) .. (170,170) -- (270,170) .. controls (275.52,170) and (280,174.48) .. (280,180) -- (280,210) .. controls (280,215.52) and (275.52,220) .. (270,220) -- (170,220) .. controls (164.48,220) and (160,215.52) .. (160,210) -- cycle ;
\draw   (10,100) .. controls (10,94.48) and (14.48,90) .. (20,90) -- (120,90) .. controls (125.52,90) and (130,94.48) .. (130,100) -- (130,130) .. controls (130,135.52) and (125.52,140) .. (120,140) -- (20,140) .. controls (14.48,140) and (10,135.52) .. (10,130) -- cycle ;
\draw   (10,340) .. controls (10,334.48) and (14.48,330) .. (20,330) -- (120,330) .. controls (125.52,330) and (130,334.48) .. (130,340) -- (130,370) .. controls (130,375.52) and (125.52,380) .. (120,380) -- (20,380) .. controls (14.48,380) and (10,375.52) .. (10,370) -- cycle ;
\draw  [color={rgb, 255:red, 245; green, 166; blue, 35 }  ,draw opacity=1 ][dash pattern={on 5.63pt off 4.5pt}][line width=1.5]  (140,340) .. controls (140,334.48) and (144.48,330) .. (150,330) -- (420,330) .. controls (425.52,330) and (430,334.48) .. (430,340) -- (430,370) .. controls (430,375.52) and (425.52,380) .. (420,380) -- (150,380) .. controls (144.48,380) and (140,375.52) .. (140,370) -- cycle ;
\draw   (10,20) .. controls (10,14.48) and (14.48,10) .. (20,10) -- (120,10) .. controls (125.52,10) and (130,14.48) .. (130,20) -- (130,50) .. controls (130,55.52) and (125.52,60) .. (120,60) -- (20,60) .. controls (14.48,60) and (10,55.52) .. (10,50) -- cycle ;
\draw   (10,260) .. controls (10,254.48) and (14.48,250) .. (20,250) -- (120,250) .. controls (125.52,250) and (130,254.48) .. (130,260) -- (130,290) .. controls (130,295.52) and (125.52,300) .. (120,300) -- (20,300) .. controls (14.48,300) and (10,295.52) .. (10,290) -- cycle ;
\draw [line width=1.5]    (284,80) -- (336,80) ;
\draw [shift={(340,80)}, rotate = 180] [fill={rgb, 255:red, 0; green, 0; blue, 0 }  ][line width=0.08]  [draw opacity=0] (13.4,-6.43) -- (0,0) -- (13.4,6.44) -- (8.9,0) -- cycle    ;
\draw [shift={(280,80)}, rotate = 0] [fill={rgb, 255:red, 0; green, 0; blue, 0 }  ][line width=0.08]  [draw opacity=0] (13.4,-6.43) -- (0,0) -- (13.4,6.44) -- (8.9,0) -- cycle    ;
\draw [line width=1.5]    (180,80) -- (216,80) ;
\draw [shift={(220,80)}, rotate = 180] [fill={rgb, 255:red, 0; green, 0; blue, 0 }  ][line width=0.08]  [draw opacity=0] (13.4,-6.43) -- (0,0) -- (13.4,6.44) -- (8.9,0) -- cycle    ;
\draw [line width=1.5]    (280,200) -- (306,200) ;
\draw [shift={(310,200)}, rotate = 180] [fill={rgb, 255:red, 0; green, 0; blue, 0 }  ][line width=0.08]  [draw opacity=0] (13.4,-6.43) -- (0,0) -- (13.4,6.44) -- (8.9,0) -- cycle    ;
\draw [line width=1.5]    (130,200) -- (156,200) ;
\draw [shift={(160,200)}, rotate = 180] [fill={rgb, 255:red, 0; green, 0; blue, 0 }  ][line width=0.08]  [draw opacity=0] (13.4,-6.43) -- (0,0) -- (13.4,6.44) -- (8.9,0) -- cycle    ;
\draw [line width=1.5]    (70,224) -- (70,246) ;
\draw [shift={(70,250)}, rotate = 270] [fill={rgb, 255:red, 0; green, 0; blue, 0 }  ][line width=0.08]  [draw opacity=0] (13.4,-6.43) -- (0,0) -- (13.4,6.44) -- (8.9,0) -- cycle    ;
\draw [shift={(70,220)}, rotate = 90] [fill={rgb, 255:red, 0; green, 0; blue, 0 }  ][line width=0.08]  [draw opacity=0] (13.4,-6.43) -- (0,0) -- (13.4,6.44) -- (8.9,0) -- cycle    ;
\draw [color={rgb, 255:red, 245; green, 166; blue, 35 }  ,draw opacity=1 ][line width=1.5]  [dash pattern={on 5.63pt off 4.5pt}]  (70,330) -- (70,303) ;
\draw [shift={(70,300)}, rotate = 90] [color={rgb, 255:red, 245; green, 166; blue, 35 }  ,draw opacity=1 ][line width=1.5]    (14.21,-6.37) .. controls (9.04,-2.99) and (4.3,-0.87) .. (0,0) .. controls (4.3,0.87) and (9.04,2.99) .. (14.21,6.37)   ;
\draw [color={rgb, 255:red, 245; green, 166; blue, 35 }  ,draw opacity=1 ][line width=1.5]  [dash pattern={on 5.63pt off 4.5pt}]  (70,60) -- (70,87) ;
\draw [shift={(70,90)}, rotate = 270] [color={rgb, 255:red, 245; green, 166; blue, 35 }  ,draw opacity=1 ][line width=1.5]    (14.21,-6.37) .. controls (9.04,-2.99) and (4.3,-0.87) .. (0,0) .. controls (4.3,0.87) and (9.04,2.99) .. (14.21,6.37)   ;
\draw [color={rgb, 255:red, 245; green, 166; blue, 35 }  ,draw opacity=1 ][line width=1.5]  [dash pattern={on 5.63pt off 4.5pt}]  (370,80) -- (417,80) ;
\draw [shift={(420,80)}, rotate = 180] [color={rgb, 255:red, 245; green, 166; blue, 35 }  ,draw opacity=1 ][line width=1.5]    (14.21,-6.37) .. controls (9.04,-2.99) and (4.3,-0.87) .. (0,0) .. controls (4.3,0.87) and (9.04,2.99) .. (14.21,6.37)   ;
\draw [color={rgb, 255:red, 245; green, 166; blue, 35 }  ,draw opacity=1 ][line width=1.5]  [dash pattern={on 5.63pt off 4.5pt}]  (70,140) -- (70,167) ;
\draw [shift={(70,170)}, rotate = 270] [color={rgb, 255:red, 245; green, 166; blue, 35 }  ,draw opacity=1 ][line width=1.5]    (14.21,-6.37) .. controls (9.04,-2.99) and (4.3,-0.87) .. (0,0) .. controls (4.3,0.87) and (9.04,2.99) .. (14.21,6.37)   ;
\draw [line width=1.5]    (134,280) -- (370,280) -- (370,220) ;
\draw [shift={(130,280)}, rotate = 0] [fill={rgb, 255:red, 0; green, 0; blue, 0 }  ][line width=0.08]  [draw opacity=0] (13.4,-6.43) -- (0,0) -- (13.4,6.44) -- (8.9,0) -- cycle    ;
\draw [color={rgb, 255:red, 245; green, 166; blue, 35 }  ,draw opacity=1 ][line width=1.5]  [dash pattern={on 5.63pt off 4.5pt}]  (130,120) -- (220,120) -- (220,167) ;
\draw [shift={(220,170)}, rotate = 270] [color={rgb, 255:red, 245; green, 166; blue, 35 }  ,draw opacity=1 ][line width=1.5]    (14.21,-6.37) .. controls (9.04,-2.99) and (4.3,-0.87) .. (0,0) .. controls (4.3,0.87) and (9.04,2.99) .. (14.21,6.37)   ;
\draw [color={rgb, 255:red, 245; green, 166; blue, 35 }  ,draw opacity=1 ][line width=1.5]  [dash pattern={on 5.63pt off 4.5pt}]  (220,120) -- (370,120) -- (370,167) ;
\draw [shift={(370,170)}, rotate = 270] [color={rgb, 255:red, 245; green, 166; blue, 35 }  ,draw opacity=1 ][line width=1.5]    (14.21,-6.37) .. controls (9.04,-2.99) and (4.3,-0.87) .. (0,0) .. controls (4.3,0.87) and (9.04,2.99) .. (14.21,6.37)   ;
\draw  [dash pattern={on 0.84pt off 2.51pt}] (140,10) -- (430,10) -- (430,110) -- (140,110) -- cycle ;

\draw (70,195) node   [align=left] {Repeatability};
\draw (220,195) node   [align=left] {Reproducibility};
\draw (370,195) node   [align=left] {Replicability};
\draw (70,355) node   [align=left] {Adaptability};
\draw (70,115) node   [align=left] {Model Selection};
\draw (70,35) node   [align=left] {Data Quality};
\draw (285,355) node   [align=left] {Meta/Incentives: influence all other parts. };
\draw (70,275) node   [align=left] {Maintainability};
\draw (310,45) node   [align=left] {\begin{minipage}[lt]{68pt}\setlength\topsep{0pt}
\begin{center}
Interact With Eachother
\end{center}

\end{minipage}};
\draw (200,45) node   [align=left] {\begin{minipage}[lt]{81.6pt}\setlength\topsep{0pt}
\begin{center}
Is a precodition for the target
\end{center}

\end{minipage}};
\draw (395,45) node   [align=left] {\begin{minipage}[lt]{47.6pt}\setlength\topsep{0pt}
\begin{center}
Strongly Influences
\end{center}

\end{minipage}};

\end{tikzpicture}
    }
    \caption{Connections on how rigor types influence each other. Solid lines indicate hard dependencies, while dashed lines show influencing effects.}
    \label{fig:whatImpactsWho}
\end{figure}
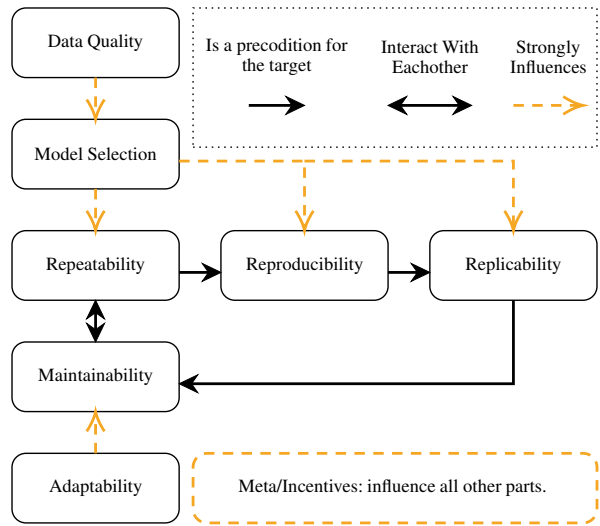

Having defined a set of eight rigor types that are being worked on, we further elaborate on our perception of connections between these rigors. In particular, there are direct and indirect relationships, which are summarized in
\autoref{fig:whatImpactsWho} with solid and dashed lines, respectively.

\subsection{Direct Relationships}

The most obvious, and intuitive connections are from repeatability to reproducibility to repeatability, as each requires a progressive step of
difficulty from the prior. If a single person/team cannot repeat their
own experiments, there is no reason to believe that a different person with
the same code would be able to reproduce those results. Extended
further, if they cannot reproduce the results with the original code,
there is no special reason to believe that by writing their own code or
using different data, they would be able to replicate the results.

Less obvious are the interactions between maintainability, repeatability, and replicability. The first is the two-way relationship between repeatability and maintainability. If an AI/ML system is not repeatable, it cannot be maintainable, as repeatability is the property that we want to maintain. Similarly, if it cannot be maintained, it may not be repeatable \emph{over time}. A simple case is the use of Docker to gain repeatability, which is predicated on the repeatability of Docker containers. This assumption is true on short time horizons, but changes in software, hardware, and eventually deprecation of tools like Docker itself do not make it true in perpetuity. The time-based evolution that maintainability requires then directly implies the replicability of a method. If a system is replicable, meaning that the code or data can change as well as the people, it satisfies the requirement of maintainability over a single point in time. Thus, maintainability involves iterated replicability over time and instantaneous repeatability at any point in time.

\subsection{Indirect Relationships}

Beyond the general influence of meta- and incentives-based rigor having a relationship to all parts of scientific rigor, we can further draw other connections that are of particular note. The most straightforward of these is that of model selection on repeatability, reproducibility, and replicability, each of which will often incorporate the model selection task as part of the motivation for why the proposed work should be used (i.e., it was demonstrated to ``be better'' than something prior). Thus, by its nature, different approaches to model selection will influence each. For example, the use of random search as a hyperparameter tuning method \citep{Bergstra2012} is potentially a hindrance to replicability due to higher variance, even if it is easily repeatable and reproducible given the original code with initial seed values for the pseudorandom number generator.

Upstream from this concern is then label and data quality, which will influence what features are selected. This is particularly notable as many datasets reach high accuracies where ``errors'' in the model's predictions are discovered to be either 1) correct and that the test data were mislabeled or 2) that the test instance was inherently ambiguous \citep{Barz2019}. This creates a new kind of noise in the selection process, and can thus alter conclusions on the merits of what is considered. This is particularly true for the eventual selection of the downstream model under replicability, where the data in use may be different.

Finally, we note that a method that is adaptable is more likely to be maintainable. The nature of one method being effective in many others is the observation that many small details on the implementation can vary, while still producing quantitatively similar results, an often observed phenomenon in decision tree literature \citep{Quinlan1993,Breiman1984,Quinlan2006,JMLR:v18:16-131}. This provides some inherent ``robustness'' to issues that often cause maintainability problems, such as changes in low-level libraries like BLAS/LAPACK or new hardware.

\section{Conclusions} \label{sec:conclusions}

We have synthesized eight current directions in the literature of scientific rigor for machine learning, disentangling them from the commonly repeated moniker of ``reproducibility'' and thus quantified the proportion of each type as studied today. These rigor types have been further characterized by their interactions/dependencies with each other. 

\bibliography{references,extraRefs}

\clearpage
\onecolumn
\appendix
\section{Papers Used} \label{sec:papers}

In the following table, we cite all the papers (101) we categorized as being significantly related to AI/ML and self-identify as being about ``reproducibility'' in some sense to build \autoref{tbl:themes}. Articles are listed in no particular order. Articles that did not self-identify as being about ``reproducibility'' are excluded as the purpose was to determine what researchers currently identify, though we found many more articles that discuss the same themes/issues in both historical and current literature (as the long bibliography demonstrates). The category assigned is our subjective call as to the most important/prominent theme of the paper, though many papers discussed more than one issue. 

\begin{longtable}{cc}
\toprule
Citation & Category \\ \midrule
\cite{Geiger2021}         &  Repeatability         \\
\cite{Musgrave2020} & Reproducibility\\ 
\cite{Tatman2018} & Reproducibility \\ 
\cite{Shepherd2017} & Reproducibility \\ 
\cite{Raff2019_quantify_repro} & Replicability \\ 
\cite{Zaharia2018AcceleratingTM} & Repeatability \\ 
\cite{Bouthillier2021} & Model Selection \\ 
\cite{Beyer2020} & Label Quality \\ 
\cite{10.1145/3383313.3412489} & Model Selection \\ 
\cite{Dacrema2019} & Model Selection \\ 
\cite{gesi_code_2022} & Maintainability \\ 
\cite{Paganini2020} & Repeatability \\ 
\cite{dror-etal-2019-deep} & Model Selection \\ 
\cite{gundersen_machine_2022} & Maintainability \\ 
\cite{Barz2019} & Label Quality \\ 
\cite{Raff2022} & Incentives \\ 
\cite{Gardner2018} & Repeatability \\ 
\cite{coakley_examining_2022} & Maintainability \\ 
\cite{Marchesin2020} & Adaptability \\ 
\cite{Eggensperger2021} & Model Selection \\ 
\cite{Cooper2021} & Model Selection \\ 
\cite{pmlr-v119-engstrom20a} & Label Quality \\ 
\cite{Publio2018} & Repeatability \\ 
\cite{matsui_mlops_2022} & Repeatability \\ 
\cite{gundersen_reproducible_2018} & Reproducibility \\ 
\cite{Liu2020e} & Reproducibility \\ 
\cite{Zhuang2021} & Maintainability \\ 
\cite{dror-etal-2017-replicability} & Model Selection \\ 
\cite{Ahn2022} & Replicability \\ 
\cite{Forde2018} & Reproducibility \\ 
\cite{Drummond2018} & Repeatability \\ 
\cite{Forde2018ReproducingML} & Reproducibility \\ 
\cite{Raff2020c} & Meta \\ 
\cite{marie-etal-2021-scientific} & Meta \\ 
\cite{Dodge2019} & Model Selection \\ 
\cite{wieling_squib_2018} & Meta \\ 
\cite{Gundersen2018} & Reproducibility \\ 
\cite{dror-etal-2018-hitchhikers} & Meta \\ 
\cite{Kim2022} & Model Selection \\ 
\cite{pmlr-v97-bouthillier19a} & Model Selection \\ 
\cite{mlsys2020_73} & Model Selection \\ 
\cite{lu_coreset_2023} & Model Selection \\ 
\cite{chen_towards_2022} & Repeatability \\ 
\cite{sun_daisyrec_2022} & Model Selection \\ 
\cite{yildiz_reproducedpapersorg_2021} & Replicability \\ 
\cite{lucic_towards_2022} & Repeatability \\ 
\cite{belz_reprogen_2020} & Label Quality \\ 
\cite{lopresti_reproducibility_2021} & Meta \\ 
\cite{moreau_benchopt_2022} & Model Selection \\ 
\cite{arpteg_software_2018} & Maintainability \\ 
\cite{tang_empirical_2021} & Maintainability \\ 
\cite{bogner_characterizing_2021} & Maintainability \\ 
\cite{10.5555/2969442.2969519} & Maintainability \\ 
\cite{kery_interactions_2018} & Repeatability \\ 
\cite{head_managing_2019} & Repeatability \\ 
\cite{yang_subtle_2021} & Maintainability \\ 
\cite{rao_where_2022} & Reproducibility \\ 
\cite{huang_state_2022} & Reproducibility \\ 
\cite{rahmani_experiments_2022} & Adaptability \\ 
\cite{roy_users_2022} & Adaptability \\ 
\cite{wang_inspection_2022} & Replicability \\ 
\cite{Bates2021} & Model Selection\\ 
\cite{Bergmeir:2018:NVC:3178572.3178665} & Model Selection\\ 
\cite{VAROQUAUX201868} & Model Selection\\ 
\cite{Soboroff:2018:MRE:3269206.3271719} & Model Selection\\ 
\cite{10.5555/3504035.3504427} & Replicability \\
\cite{pmlr-v68-johnson17a} & Replicability \\
\cite{pmlr-v85-hegselmann18a} & Replicability \\
\cite{McDermott_Wang_Marinsek_Ranganath_Foschini_Ghassemi_2021} & Meta \\
\cite{8621874} & Repeatability \\
\cite{lin-etal-2023-linear} & Reproducibility \\
\cite{liu-etal-2021-parameter} & Adaptability \\
\cite{chen-etal-2022-even} & Replicability \\ 
\cite{Lin_Liu_Chen_Hsu_Wu_Tsai_Lin_2022} & Reproducibility \\
\cite{pmlr-v202-kalavasis23a} & Replicability \\
\cite{karbasi2023replicability} & Replicability \\
\cite{esfandiari2023replicable} & Replicability \\
\cite{zhang2023optimal} & Replicability \\
\cite{esfandiari2023replicableBandits} & Replicability \\
\cite{10.1145/3564246.3585246} & Replicability \\ 
\cite{10.1145/3519935.3519973} & Replicability \\
\cite{KAPOOR2023100804} & Meta \\
\cite{10.1145/3514094.3534196} & Meta \\
\cite{10.1145/3576915.3623130} & Meta \\
\cite{arvan_reproducibility_2022} & Meta \\
\cite{storks-etal-2023-nlp} & Meta \\
\cite{arvan23_interspeech} & Meta \\
\cite{10.1145/3596519} & Meta \\ 
\cite{raff2023reproducibility} & Reproducibility \\
\cite{belz-etal-2022-quantified} &  Meta \\
\cite{papi-etal-2024-good} & Maintainability  \\
\cite{belz-etal-2022-2022} & Replicability  \\
\cite{Inel2023} & Maintainability  \\
\cite{chen-etal-2022-reproducibility} &  Reproducibility \\
\cite{ito-etal-2023-challenges} &  Replicability \\
\cite{Connolly2023Software} &  Maintainability \\
\cite{10.1145/3641525.3663628} &  Meta \\
\cite{10.1145/3641525.3663622} &  Repeatability \\
\cite{10.1145/3641525.3663619} & Model Selection  \\
\cite{10.1145/3641525.3663618} &  Model Selection \\
\cite{NEURIPS2023_af076c3b} & Repeatability \\
\bottomrule
\end{longtable}

\end{document}